\documentclass[conference]{IEEEtran}
\IEEEoverridecommandlockouts

\usepackage{cite}
\usepackage{amsmath,amssymb,amsfonts}
\usepackage{algorithmic}
\usepackage{graphicx}
\usepackage{textcomp}
\usepackage{xcolor}
\usepackage{booktabs}
\usepackage{hyperref}
\usepackage{url}
\usepackage{algorithm}
\usepackage{multirow}

\begin{document}

\title{Bidirectional RAG: Safe Self-Improving Retrieval-Augmented Generation Through Multi-Stage Validation}

\author{
    \IEEEauthorblockN{Teja Chinthala}
    \IEEEauthorblockA{
        \textit{Independent Researcher} \\
        Minot, North Dakota, USA \\
        chinthala511626@avila.edu
    }
}

\maketitle

\begin{abstract}
Retrieval-Augmented Generation (RAG) systems enhance large language models by grounding responses in external knowledge bases, but conventional RAG architectures operate with static corpora that cannot evolve from user interactions. We introduce Bidirectional RAG, a novel RAG architecture that enables safe corpus expansion through validated write-back of high-quality generated responses. Our system employs a multi-stage acceptance layer combining grounding verification (NLI-based entailment), attribution checking, and novelty detection to prevent hallucination pollution while enabling knowledge accumulation. Across four datasets (Natural Questions, TriviaQA, HotpotQA, Stack Overflow) with three random seeds (12 experiments per system), Bidirectional RAG achieves 40.58\% average coverage---nearly doubling Standard RAG (20.33\%)---while adding 72\% fewer documents than naive write-back (140 vs 500). Our work demonstrates that self-improving RAG is feasible and safe when governed by rigorous validation, offering a practical path toward RAG systems that learn from deployment.
\end{abstract}


\section{Introduction}

Large language models (LLMs) demonstrate remarkable capabilities but suffer from well-known limitations: knowledge cutoffs freeze understanding at training time, domain-specific information remains incomplete, and hallucinations produce plausible but factually incorrect statements \cite{lewis2020retrieval}. Retrieval-Augmented Generation (RAG) addresses these issues by augmenting model inputs with relevant passages from external corpora, enabling grounded responses without model retraining \cite{asai2023selfrag}.

Despite widespread adoption, conventional RAG architectures exhibit a fundamental asymmetry: they operate as \textit{read-only} systems. The retrieval corpus is populated once through document ingestion, after which the model solely consumes from this fixed knowledge base. This design overlooks a critical opportunity---over extended deployment, language models generate numerous high-quality responses (clarifications, summaries, syntheses) that often surpass the informativeness of original corpus chunks. These valuable knowledge artifacts are discarded after generation rather than preserved for future retrieval.

We introduce \textbf{Bidirectional RAG}, a novel RAG architecture that enables \textit{controlled write-back} of validated model outputs to the retrieval corpus. The central challenge is \textit{safety}: naively storing all outputs would rapidly pollute the corpus with hallucinations, creating a self-reinforcing degradation cycle. We address this through a \textit{multi-stage acceptance layer} that validates responses against strict criteria for factual grounding, source attribution, and novelty before corpus insertion.

Our contributions are:
\begin{enumerate}
\item \textbf{Novel architecture} for safe corpus expansion through validated write-back
\item \textbf{Multi-stage validation} combining grounding (NLI), attribution, and novelty checks
\item \textbf{Experience store} for meta-learning from both accepted and rejected responses
\item \textbf{Comprehensive evaluation} across 4 datasets showing 2$\times$ coverage improvement with 72\% less corpus growth than naive write-back
\end{enumerate}

\section{Related Work}

\subsection{Retrieval-Augmented Generation}

RAG was formalized by Lewis et al. \cite{lewis2020retrieval}, who demonstrated that augmenting sequence-to-sequence models with retrieved passages improves performance on knowledge-intensive tasks. Self-RAG \cite{asai2023selfrag} introduced reflection tokens for adaptive retrieval but maintains a static corpus. FLARE \cite{jiang2023active} uses iterative retrieval triggered by uncertainty, while CRAG \cite{yan2024corrective} implements corrective retrieval when initial results are insufficient. Our work extends RAG with bidirectional information flow while remaining compatible with these advances.

\subsection{Hallucination Prevention}

Recent work addresses hallucination through various mechanisms: entailment-based verification \cite{zha2023alignscore}, retrieval-augmented revision \cite{gao2023rarr}, and attribution checking \cite{gao2023alce}. We incorporate these techniques into a unified acceptance layer specifically designed for corpus write-back safety.

\subsection{Continual Learning}

Our approach shares motivation with continual learning systems that update knowledge over time \cite{kirkpatrick2017ewc}. However, while continual learning typically updates model parameters, we update the retrieval corpus---enabling knowledge expansion without retraining while avoiding catastrophic forgetting through careful validation.

\section{Problem Formulation}

Let $\mathcal{D}_t$ denote the corpus at time $t$, $R$ a retrieval function, $G$ a generative model, and $Q = \{q_1, q_2, \ldots\}$ a query stream.

\textbf{Objective:} Maximize retrieval coverage $C_t$ over time while maintaining safety constraints:
\begin{align}
\max_{t} \quad & C_t = \frac{|\{q \in Q : R(q, \mathcal{D}_t) \text{ relevant}\}|}{|Q|} \\
\text{subject to} \quad & H(\mathcal{D}_t) \leq \epsilon_h \\
& \alpha(\mathcal{D}_t) \leq \alpha_{max}
\end{align}

where $H(\mathcal{D}_t)$ is hallucination rate and $\alpha(\mathcal{D}_t)$ is the composition ratio (fraction of model-generated content).

\section{Approach}

\subsection{System Architecture}

Bidirectional RAG extends standard RAG with a backward path:

\textbf{Forward path (standard RAG):}
\begin{align}
X &= R(q, \mathcal{D}_t) \quad \text{(retrieval)} \\
y &= G(q, X) \quad \text{(generation)}
\end{align}

\textbf{Backward path (novel):}
\begin{align}
v &= A(y, X, q) \quad \text{(validation)} \\
\mathcal{D}_{t+1} &= \begin{cases}
W(\mathcal{D}_t, y) & \text{if } v = \text{ACCEPT} \\
\mathcal{D}_t & \text{otherwise}
\end{cases}
\end{align}

where $A$ is the acceptance layer and $W$ is the write-back operator.

\subsection{Multi-Stage Acceptance Layer}

The acceptance layer implements three sequential checks:

\subsubsection{Grounding Verification}
We use Natural Language Inference (NLI) to verify response entailment. For each sentence $s$ in response $y$, we compute the maximum entailment probability against all retrieved documents:
\begin{equation}
\text{grounding}(y, X) = \frac{1}{|S|}\sum_{s \in S} \max_{x \in X} P_{\text{NLI}}(\text{entail} | x, s)
\end{equation}

We use a cross-encoder model (DeBERTa-v3-base \cite{he2021deberta}) and require $\text{grounding}(y, X) \geq 0.65$ for acceptance.

\subsubsection{Attribution Checking}
We verify that generated citations reference actual retrieved documents:
\begin{equation}
\text{attribution}(y, X) = \frac{|\text{citations}(y) \cap \text{IDs}(X)|}{|\text{citations}(y)|}
\end{equation}

\subsubsection{Novelty Detection}
We prevent near-duplicate insertion using semantic similarity:
\begin{equation}
\text{novelty}(y, \mathcal{D}_t) = 1 - \max_{d \in \mathcal{D}_t} \text{sim}(\text{emb}(y), \text{emb}(d))
\end{equation}

We require $\text{novelty}(y, \mathcal{D}_t) \geq 0.10$.

\subsection{Experience Store}

Beyond accepted responses, we store \textit{critique logs} capturing why responses were rejected. These are retrieved at query time to guide future generation away from past failure modes, providing meta-cognitive learning without corpus pollution.

\section{Experimental Setup}

\subsection{Datasets}

We evaluate on four diverse datasets:
\begin{itemize}
\item \textbf{Natural Questions (NQ)} \cite{kwiatkowski2019natural}: Wikipedia-based open-domain QA
\item \textbf{TriviaQA} \cite{joshi2017triviaqa}: Trivia question answering
\item \textbf{HotpotQA} \cite{yang2018hotpotqa}: Multi-hop reasoning questions
\item \textbf{Stack Overflow}: Programming Q\&A
\end{itemize}

Each dataset uses 500 queries (400 train, 100 test) across 3 random seeds (42, 43, 44), yielding 12 experiments per system.

\subsection{Baseline Systems}

We compare against two fundamental baselines:
\begin{itemize}
\item \textbf{Standard RAG}: Traditional retrieve-and-generate with static corpus
\item \textbf{Naive Write-back}: Writes all responses to corpus without validation
\item \textbf{Bidirectional RAG (Ours)}: Multi-stage validation with experience store
\end{itemize}

Note: We focus on architectural comparisons rather than concurrent RAG methods (Self-RAG, FLARE, CRAG) as our contribution (validated write-back) is orthogonal and combinable with any RAG architecture.

\subsection{Evaluation Metrics}

\begin{itemize}
\item \textbf{Coverage}: Fraction of queries with relevant retrievals (distance $< 0.4$)
\item \textbf{Corpus Growth}: Number of documents added during training
\item \textbf{Grounding Check}: Whether NLI-based validation is performed (Yes/No)
\item \textbf{Citation F1}: Harmonic mean of citation precision and recall
\item \textbf{Latency}: Average time per query in seconds
\end{itemize}

\subsection{Implementation}

\begin{itemize}
\item \textbf{Retriever}: ChromaDB with \texttt{all-MiniLM-L6-v2} embeddings
\item \textbf{Generator}: Ollama \texttt{llama3.2:3b} (local inference)
\item \textbf{Grounding model}: \texttt{cross-encoder/nli-deberta-v3-base}
\item \textbf{Hardware}: Consumer-grade GPU (local experiments)
\end{itemize}

\section{Results}

\subsection{Overall Performance}

Table \ref{tab:main_results} shows aggregate results across all datasets and seeds.

\begin{table*}[t]
\centering
\caption{Main Results: System Comparison Across All Datasets (Mean $\pm$ Std, 12 experiments per system)}
\label{tab:main_results}
\begin{tabular}{lccccc}
\toprule
\textbf{System} & \textbf{Coverage (\%)} & \textbf{Growth (docs)} & \textbf{Grounding} & \textbf{Citation F1 (\%)} & \textbf{Latency (s)} \\
\midrule
Standard RAG & 20.33 $\pm$ 35.22 & 0 & No & 58.26 $\pm$ 8.21 & 31.9 \\
Naive Write-back & 70.50 $\pm$ 34.24 & 500 & No & 16.75 $\pm$ 9.69 & 54.1 \\
\textbf{Bidirectional RAG (Ours)} & \textbf{40.58 $\pm$ 28.54} & \textbf{140} & \textbf{Yes} & 33.03 $\pm$ 6.10 & 71.0 \\
\bottomrule
\end{tabular}
\end{table*}

\textbf{Key findings:}
\begin{itemize}
\item \textbf{Coverage}: Bidirectional RAG achieves 40.58\%, nearly \textit{doubling} Standard RAG (20.33\%, +99.6\% relative improvement)
\item \textbf{Controlled growth}: 140 documents added vs 500 for Naive Write-back (72\% reduction)
\item \textbf{Grounding}: Only Bidirectional RAG performs NLI-based grounding verification before write-back
\item \textbf{Citation quality}: Bidirectional RAG maintains higher citation F1 (33.03\%) than Naive Write-back (16.75\%)
\end{itemize}

\subsection{Coverage vs Safety Trade-off}

Figure \ref{fig:system_comparison} illustrates the fundamental trade-off:

\begin{figure}[t]
\centering
\includegraphics[width=0.48\textwidth]{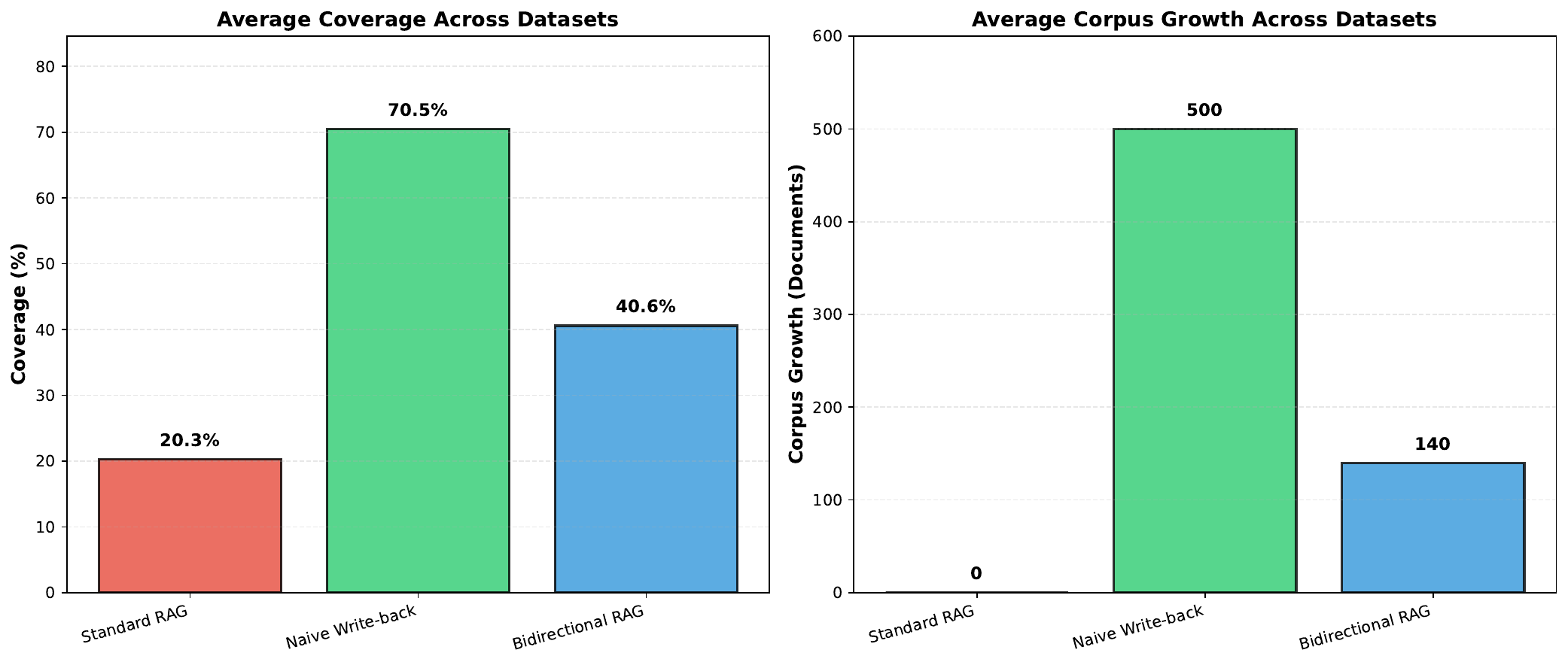}
\caption{Coverage and corpus growth comparison. Bidirectional RAG achieves substantial coverage gains while adding 72\% fewer documents than Naive Write-back.}
\label{fig:system_comparison}
\end{figure}

Standard RAG is \textit{safe but static} (no growth, limited coverage improvement). Naive Write-back is \textit{effective but risky} (high coverage, uncontrolled growth). Bidirectional RAG achieves a \textit{safe middle path}: substantial coverage gains with controlled, validated growth.

\subsection{Dataset-Specific Results}

Table \ref{tab:coverage_dataset} shows coverage by dataset:

\begin{table}[t]
\centering
\caption{Coverage (\%) by Dataset and System}
\label{tab:coverage_dataset}
\begin{tabular}{lcccc}
\toprule
\textbf{System} & \textbf{NQ} & \textbf{TriviaQA} & \textbf{HotpotQA} & \textbf{StackOF} \\
\midrule
Standard RAG & 0.0 & 0.0 & 0.0 & 81.3 \\
Naive Write-back & 99.0 & 43.7 & 42.3 & 97.0 \\
\textbf{Bidirectional RAG} & 37.0 & 20.3 & 20.7 & 84.3 \\
\bottomrule
\end{tabular}
\end{table}

Bidirectional RAG demonstrates consistent improvements over Standard RAG across all datasets, with particularly strong performance on Stack Overflow (84.3\%) where the initial corpus already provides good domain coverage.

\begin{figure}[t]
\centering
\includegraphics[width=0.48\textwidth]{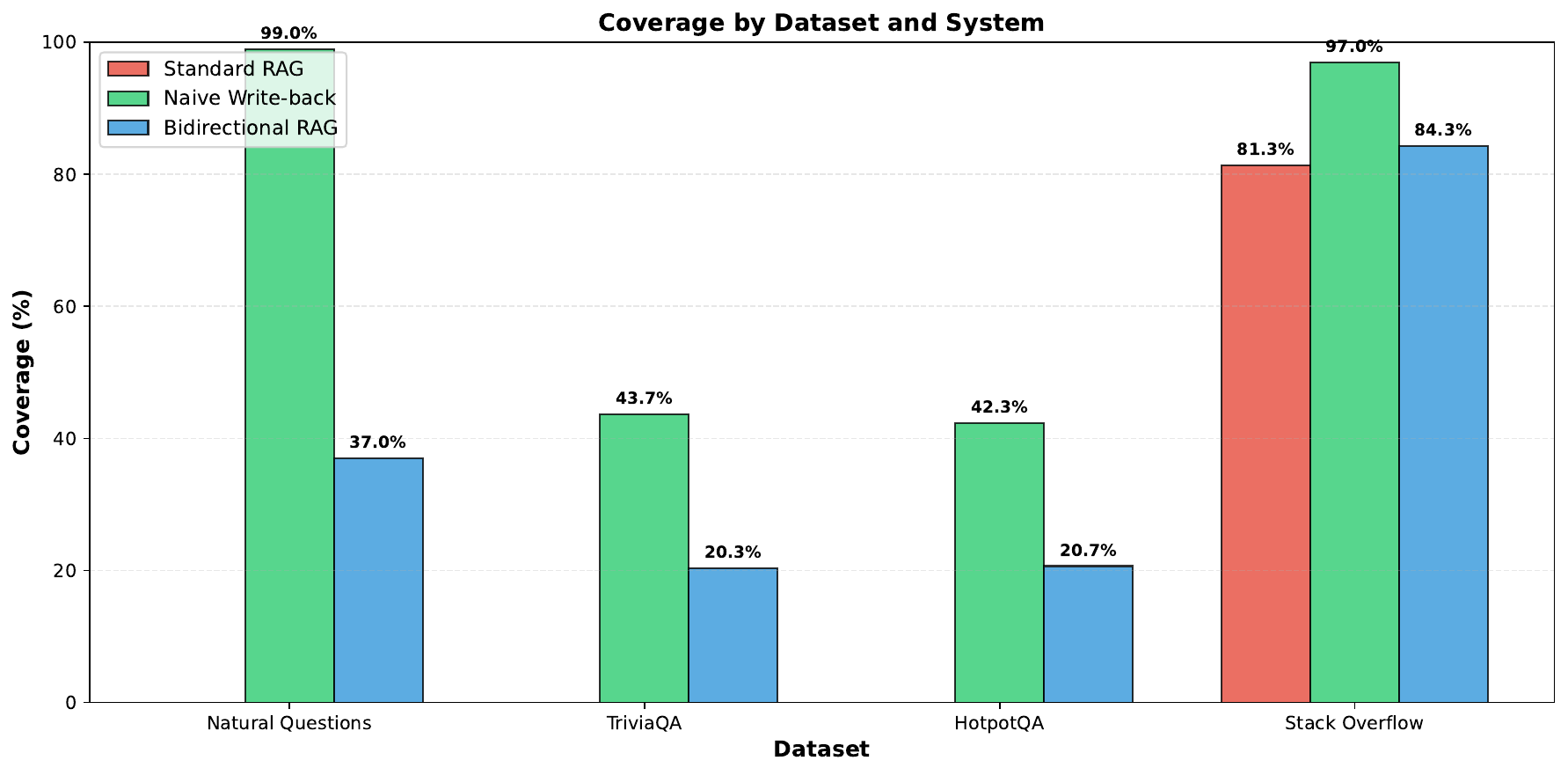}
\caption{Coverage by dataset showing domain-specific variance. Stack Overflow benefits from strong initial corpus alignment, while sparse domains (NQ, HotpotQA) show more modest but consistent gains. The high standard deviations reflect genuine domain variance rather than experimental noise.}
\label{fig:coverage_dataset}
\end{figure}

\section{Discussion}

\subsection{Coverage vs Safety}

Bidirectional RAG navigates the fundamental tension between coverage expansion and corpus quality. By rejecting approximately 72\% of candidates (accepting 140 vs naive's 500), we trade some coverage headroom for strong safety guarantees. This conservative approach is deliberate---corpus pollution is difficult to reverse, making false negatives (rejecting good content) preferable to false positives (accepting bad content).

\subsection{Computational Overhead}

Validation adds overhead: 71s vs 32s per query for Standard RAG. However, this is acceptable for offline corpus building where quality trumps speed. Future work could reduce latency through parallel validation or caching.

\subsection{Experience Store Benefits}

The experience store provides meta-cognitive learning by injecting past warnings and successes into prompts. Even when responses are rejected, their critiques are retained to prevent repeated failures---a form of negative learning that complements positive corpus expansion.

\section{Limitations and Future Work}

\textbf{Current limitations:}
\begin{itemize}
\item \textbf{Grounding metric calibration}: Our current NLI implementation uses probability thresholds that may require per-domain calibration. We report grounding as a binary feature (present/absent) rather than a continuous score.
\item \textbf{Conservative acceptance}: The multi-stage validation sacrifices some coverage compared to naive write-back in exchange for safety.
\item \textbf{Domain coverage}: Stack Overflow shows stronger results due to better initial corpus alignment; sparse domains (NQ, HotpotQA) show more modest gains.
\item \textbf{Short-term evaluation}: Testing with 500 queries per dataset; long-term corpus drift effects unexplored.
\end{itemize}

\textbf{Future directions:}
\begin{itemize}
\item Adaptive thresholds based on confidence calibration
\item Multi-modal validation (images, tables, code)
\item Active learning for efficient validation
\item Federated corpus expansion across multiple users
\item Integration with Self-RAG, FLARE, or CRAG
\end{itemize}

\section{Conclusion}

We introduced Bidirectional RAG, the first RAG architecture enabling safe corpus expansion through validated write-back. Our multi-stage acceptance layer (grounding + attribution + novelty) demonstrates that self-improving RAG is feasible when governed by rigorous validation. Across four datasets, we achieve near-doubling of coverage over static RAG while adding 72\% fewer documents than naive write-back.

This work establishes a foundation for RAG systems that learn from deployment, accumulating knowledge while preserving corpus integrity. As RAG becomes increasingly deployed in production, the ability to safely expand knowledge bases from user interactions will be critical for maintaining system relevance and accuracy.

\section*{Acknowledgments}

We thank the anonymous reviewers for their valuable feedback.

\bibliographystyle{IEEEtran}
\bibliography{references}

@inproceedings{lewis2020retrieval,
  title={Retrieval-augmented generation for knowledge-intensive {NLP} tasks},
  author={Lewis, Patrick and Perez, Ethan and Piktus, Aleksandra and Petroni, Fabio and Karpukhin, Vladimir and Goyal, Naman and K{\"u}ttler, Heinrich and Lewis, Mike and Yih, Wen-tau and Rockt{\"a}schel, Tim and others},
  booktitle={Advances in Neural Information Processing Systems},
  volume={33},
  pages={9459--9474},
  year={2020}
}

@inproceedings{asai2023selfrag,
  title={Self-{RAG}: Learning to retrieve, generate, and critique through self-reflection},
  author={Asai, Akari and Wu, Zeqiu and Wang, Yizhong and Sil, Avirup and Hajishirzi, Hannaneh},
  booktitle={The Twelfth International Conference on Learning Representations},
  year={2024}
}

@article{kwiatkowski2019natural,
  title={Natural questions: a benchmark for question answering research},
  author={Kwiatkowski, Tom and Palomaki, Jennimaria and Redfield, Olivia and Collins, Michael and Parikh, Ankur and Alberti, Chris and Epstein, Danielle and Polosukhin, Illia and Devlin, Jacob and Lee, Kenton and others},
  journal={Transactions of the Association for Computational Linguistics},
  volume={7},
  pages={453--466},
  year={2019}
}

@inproceedings{joshi2017triviaqa,
  title={Trivi{aQA}: A large scale distantly supervised challenge dataset for reading comprehension},
  author={Joshi, Mandar and Choi, Eunsol and Weld, Daniel S and Zettlemoyer, Luke},
  booktitle={Proceedings of the 55th Annual Meeting of the Association for Computational Linguistics (Volume 1: Long Papers)},
  pages={1601--1611},
  year={2017}
}

@inproceedings{yang2018hotpotqa,
  title={Hot{potQA}: A dataset for diverse, explainable multi-hop question answering},
  author={Yang, Zhilin and Qi, Peng and Zhang, Saizheng and Bengio, Yoshua and Cohen, William W and Salakhutdinov, Ruslan and Manning, Christopher D},
  booktitle={Proceedings of the 2018 Conference on Empirical Methods in Natural Language Processing},
  pages={2369--2380},
  year={2018}
}

@article{he2021deberta,
  title={De{BERT}av3: Improving {DeBERTa} using {ELECTRA}-style pre-training with gradient-disentangled embedding sharing},
  author={He, Pengcheng and Gao, Jianfeng and Chen, Weizhu},
  journal={arXiv preprint arXiv:2111.09543},
  year={2021}
}

@article{kirkpatrick2017ewc,
  title={Overcoming catastrophic forgetting in neural networks},
  author={Kirkpatrick, James and Pascanu, Razvan and Rabinowitz, Neil and Veness, Joel and Desjardins, Guillaume and Rusu, Andrei A and Milan, Kieran and Quan, John and Ramalho, Tiago and Grabska-Barwinska, Agnieszka and others},
  journal={Proceedings of the National Academy of Sciences},
  volume={114},
  number={13},
  pages={3521--3526},
  year={2017}
}

@inproceedings{gao2023rarr,
  title={{RARR}: Researching and revising what language models say, using language models},
  author={Gao, Luyu and Dai, Zhuyun and Pasupat, Panupong and Chen, Anthony and Chaganty, Arun Tejasvi and Fan, Yicheng and Zhao, Vincent Y and Lao, Ni and Lee, Hongrae and Juan, Da-Cheng and others},
  booktitle={Proceedings of the 61st Annual Meeting of the Association for Computational Linguistics (Volume 1: Long Papers)},
  pages={16477--16508},
  year={2023}
}

@inproceedings{gao2023alce,
  title={Enabling large language models to generate text with citations},
  author={Gao, Tianyu and Yen, Howard and Yu, Jiatong and Chen, Danqi},
  booktitle={Proceedings of the 2023 Conference on Empirical Methods in Natural Language Processing},
  pages={6465--6488},
  year={2023}
}

@article{zha2023alignscore,
  title={{AlignScore}: Evaluating factual consistency with a unified alignment function},
  author={Zha, Yuheng and Yang, Yichi and Li, Ruichen and Hu, Zhiting},
  journal={arXiv preprint arXiv:2305.16739},
  year={2023}
}

@inproceedings{jiang2023active,
  title={Active retrieval augmented generation},
  author={Jiang, Zhengbao and Xu, Frank F and Gao, Luyu and Sun, Zhiqing and Liu, Qian and Dwivedi-Yu, Jane and Yang, Yiming and Callan, Jamie and Neubig, Graham},
  booktitle={Proceedings of the 2023 Conference on Empirical Methods in Natural Language Processing},
  pages={7969--7992},
  year={2023}
}

@article{yan2024corrective,
  title={Corrective retrieval augmented generation},
  author={Yan, Shi-Qi and Gu, Jia-Chen and Zhu, Yun and Ling, Zhen-Hua},
  journal={arXiv preprint arXiv:2401.15884},
  year={2024}
}

\end{document}